# A Comparative Study of Human thermal face recognition based on Haar wavelet transform (HWT) and Local Binary Pattern (LBP)

Ayan Seal, Suranjan Ganguly, Debotosh Bhattacharjee, Mita Nasipuri, Dipak Kumar Basu

*Department of Computer Science and Engineering, Jadavpur University, Kolkata- 700032, India*

**Abstract**

Thermal infra-red (IR) images focus on changes of temperature distribution on facial muscles and blood vessels. These temperature changes can be regarded as texture features of images. A comparative study of face recognition methods working in thermal spectrum is carried out in this paper. In these study two local-matching methods based on Haar wavelet transform and Local Binary Pattern (LBP) are analyzed. Wavelet transform is a good tool to analyze multi-scale, multi-direction changes of texture. Local binary patterns (LBP) are a type of feature used for classification in computer vision. Firstly, human thermal IR face image is preprocessed and cropped the face region only from the entire image. Secondly, two different approaches are used to extract the features from the cropped face region. In the first approach, the training images and the test images are processed with Haar wavelet transform and the LL band and the average of LH/HL/HH bands sub-images are created for each face image. Then a total confidence matrix is formed for each face image by taking a weighted sum of the corresponding pixel values of the LL band and average band. For LBP feature extraction, each of the face images in training and test datasets is divided into 161 numbers of sub images, each of size 8×8 pixels. For each such sub images, LBP features are extracted which are concatenated in row wise manner. PCA is performed separately on the individual feature set for dimensionality reeducation. Finally two different classifiers are used to classify face images. One such classifier multi-layer feed forward neural network and another classifier is minimum distance classifier. The Experiments have been performed on the database created at our own laboratory and Terravic Facial IR Database.

## 1. Introduction

In the modern society there is an increasing need to track and recognize persons automatically in various areas such as in the areas of surveillance, Closed Circuit Television (CCTV) control, user authentication, Human Computer Interface (HCI), daily attendance register, airport security checks and immigration checks [1], [2], [3]. Such requirement for reliable personal identification in computerized access control has resulted in an increased interest in biometrics. The key element of biometric technology is its ability to identify a human being and enforce security. Nearly all-biometric systems work in the same manner. First, a person is registered into a database using a specified method. Information about a certain characteristic of the human is captured. This information is usually placed through an algorithm that turns the information into a code that the database stores. When the person needs to be identified, the system will take the information about the person again, translates this new information with the algorithm, and then compare the new code with the stored ones in the database to find out a possible match. Biometrics use physical characteristics or personal traits to identify a person. Physical feature is suitable for identity purpose and generally obtained from living human body. Commonly used physical features are fingerprints, facial features, hand geometry, and eye features (iris and retina) etc. So, biometrics involve using the different parts of the body. Personal trait is sometimes more appropriate for some applications which need direct physical interaction. The most commonly used personal traits are signature and voices etc. Among many biometric security systems, face recognition has drawn significant attention of the researchers for the last three decades because of its potential applications in security system. There are a number of reasons to choose face recognition for designing efficient biometric security systems. The most important one is that no physical interaction is needed. This is helpful for the cases where touching is prohibited due to hygienic reasons or religious or cultural traditions. Most of the research works in this area have focused on visible spectrum imaging due to easy availability of low cost visible band optical cameras. But, it requires an external source of illumination. Even with a considerable success for automatic face recognition techniques in many practical applications, the task of face recognition based only on the visible spectrum is still a challenging problem under uncontrolled environments. The challenges are even more philosophical when one considers the large variations in the visual stimulus due to illumination conditions, poses [4], facial expressions, aging, and disguises such as facial hair, glasses, or cosmetics. Performance of visual face recognition is sensitive to variations in illumination conditions and usually degrades significantly when the lighting is not bright or when it is not illuminating the face uniformly. The changes caused by illumination on the same individual are often larger than the differences between individuals. Various algorithms (e.g. histogram equalization, eigenfaces etc.) for compensating such variations have been studied with partial success. These techniques try to reduce the within-class variability introduced by changes in illumination. To overcome this limitation, several solutions have been designed. One solution is using 3D data obtained from 3D vision device. Such systems are less dependent on illumination changes, but they have some disadvantages: the cost of such system is high and their processing speed is low. Thermal IR images [5] have been suggested as a possible alternative in handling situations where there is no control over illumination. The wavelength ranges of different infrared spectrums are shown in Table 1.

Table 1. Wavelength ranges for different infrared spectrums

| Spectrum | Wavelength range |
|---|---|
| Visible Spectrum | 0.4-0.7 μm (micro meter / micron) |
| Near Infrared (NIR) | 0.7-1.0 μm (micro meter / micron) |
| Short-wave Infrared (SWIR) | 1-3 μm (micro meter / micron) |
| Mid-wave Infrared (MWIR) | 3-5 μm (micro meter / micron) |
| Thermal Infrared (TIR) | 8-14 μm (micro meter / micron) |

Thermal IR band is more popular to the researchers working with thermal images. Recently researchers have been using Near-IR imaging cameras for face recognition with better results [6], but SWIR and MWIR have not been used significantly till now. Thermal IR images represent the heat patterns emitted from an object and they don't consider the reflected energy. Objects emit different amounts of IR energy according to their body temperature and characteristics. Previously Thermal IR camera was costly but recently the cost of such cameras has come down considerably with the development of CCD technology [7]. Thermal images can be captured under different lighting conditions, even under completely dark environment. Using thermal images, the tasks of face detection, localization, and segmentation are comparatively easier and more reliable than those in visible band images [8]. Humans are homoeothermic and hence capable of maintaining constant temperature under different surrounding temperature and since blood vessels transport warm blood throughout the body, the thermal patterns of faces are derived primarily from the pattern of blood vessels under the skin. The vein and tissue structure of the face is unique for each human being [9], and therefore the IR images are also unique. It is known that even identical twins have different thermal patterns. An infrared camera with good sensitivity can capture images of superficial blood vessels on the human face [10] without any physical interaction. However, it has been indicated by Guyton and Hall [11] that the average diameter of blood vessels is around 10-15 μm, which is too small to be detected by current IR cameras because of the limitation in spatial resolution. The skin just above a blood vessel is on an average 0.1 °C warmer than the adjacent skin, which is beyond the thermal accuracy of current IR cameras. However the convective heat transfer effect from the flow of "hot" arterial blood in superficial vessels creates characteristic thermal imprints, which are at a gradient with the surrounding tissue. Face recognition based on thermal IR spectrum utilizes the anatomical information [12] of human face as features unique to each individual while sacrificing color recognition. Therefore, the infrared image recognition should focus on thermal distribution patterns on the facial muscles and blood vessels, mainly on cheek, forehead and nasal tip. These regional thermal distribution patterns can be regarded as the texture pattern unique for a particular face. Wavelet transform can be used to detect the multi-scale, multidirectional changes of texture. Local binary patterns (LBP) are also a well-known texture descriptor and also a successful local descriptor for face recognition under local illumination variations. Therefore this paper describes a comparative study of different approach of thermal IR human face recognition system is proposed. The paper is organized as follows: Section 2 presents about the outline proposed system. In section 3 the comparative analyses of these methods in the database created at our own laboratory and Terravic Facial IR Database are presented. Finally, in Section 4 results are discussed, and conclusions are given.

## 2. Outline of the proposed system

The proposed Thermal Face Recognition System (TFRS) can be subdivided into four main parts, namely image acquisition, image preprocessing, feature extraction and classification. The image preprocessing part involves binarization of the acquired thermal face image, extraction of largest component as the face region, finding the centroid of the face region and finally cropping of the face region in elliptic shape. The two different features extraction techniques have been discussed in this paper. The first one is to find LL band and HL/LH/HH average band images using Haar wavelet transform and the total confidence matrix is used as a feature vector. The eigen space projection is performed on feature vector to reduce the dimensionality. This reduced feature vector is fed into a classifier. The second method of features extraction technique is Local Binary Pattern (LBP). As a classifier, a back propagation feed forward neural network or a minimum distance classifier is used in this paper. The block diagram of the proposed system is given in Fig. 1. The system starts with acquisition of thermal face image and end with successful classification. The set of image processing and classification techniques which has been used here are discussed in detail in subsequent subsections.

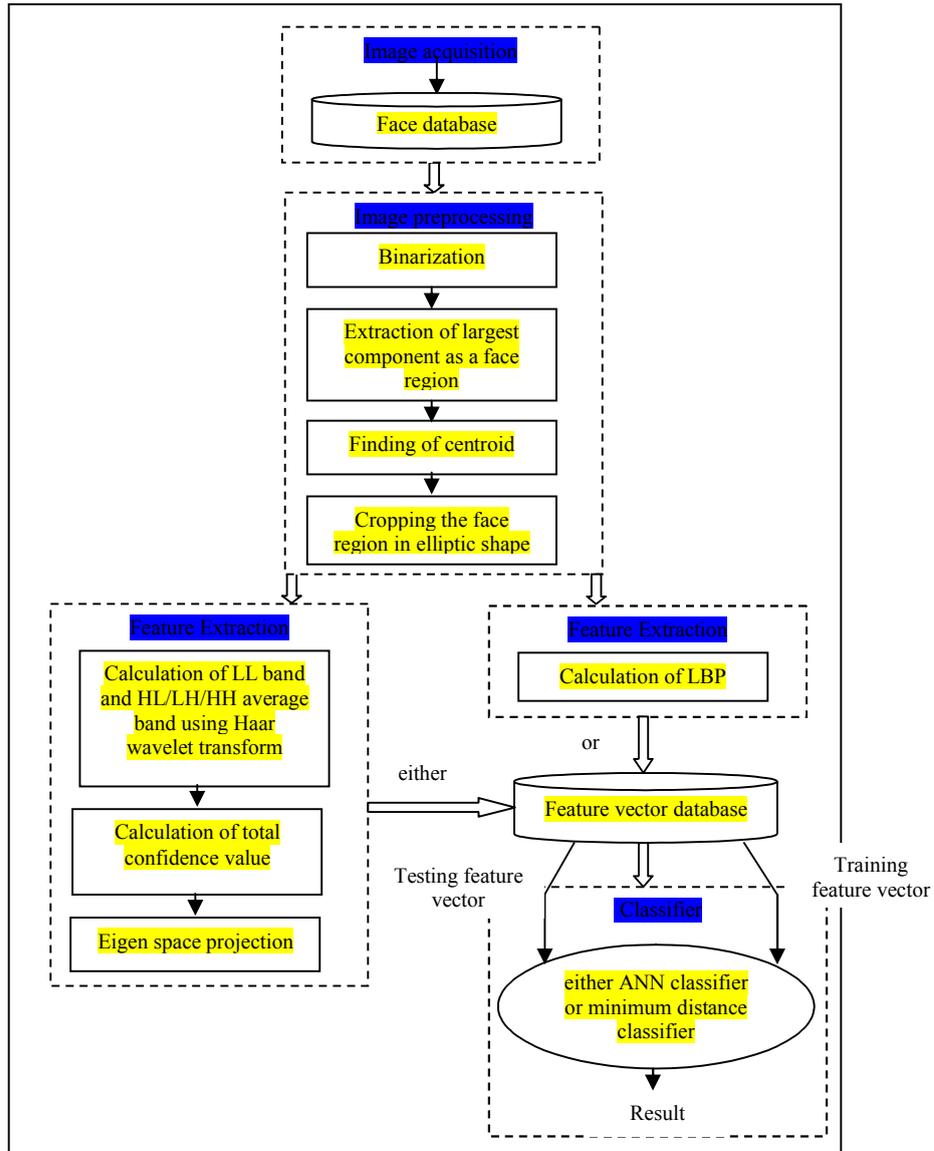

Fig. 1 Schematic block diagram of the proposed system

*2.1. Thermal face image acquisition*

In the present work unregistered thermal and visible face images are acquired simultaneously with variable expressions, poses and with/without glasses. Till now 17 individuals, have volunteered for this photo shots and for each individual 34 different templates of RGB color images with different expressions, namely (Exp1) happy, (Exp2) angry, (Exp3) sad, (Exp4) disgusted, (Exp5) neutral, (Exp6) fearful and (Exp7) surprised are available. Different pose changes about x-axis, y-axis and z-axis are also available. Resolution of each image is 320 x 240 and the images are saved in 24-bit JPEG format. Two different cameras are used to capture this database. One is Thermal – FLIR 7 and another is Optical – Sony cyber shot. A typical thermal face image is shown in Fig. 2a). This thermal face image depicts interesting thermal information of a facial model.

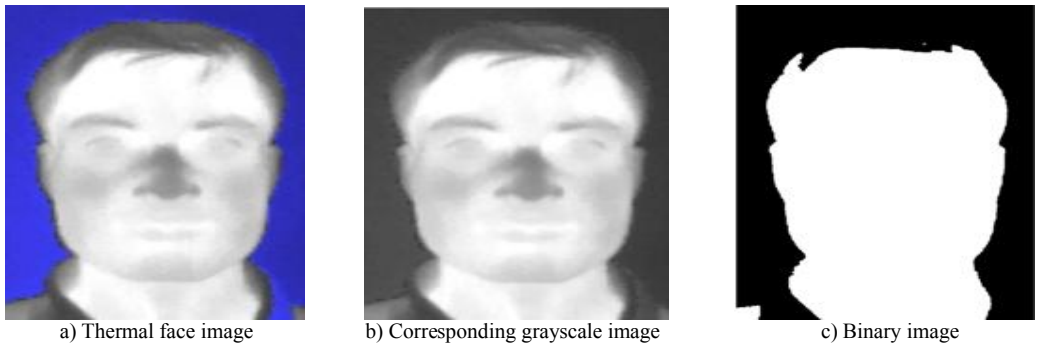

    a) Thermal face image        b) Corresponding grayscale image        c) Binary image

Fig. 2 Thermal face image and its various preprocessing stages

## 2.2. Binarization

The binarization of 24-bit colour image is divided into two steps. In first step, the colour image is converted into an 8-bit grayscale image using equation 1.

$$I = (0.2989 \times red_{component}) + (0.5870 \times green_{component}) + (0.1140 \times blue_{component}) \quad \ldots \ldots (1)$$

Where 'I' is the grayscale image. The grayscale image corresponding to the thermal image of Fig. 2a) is shown in Fig. 2b). Grayscale image is then converted into binary image. For this purpose, mean gray value of grayscale image (say $g_{mean}$) is computed with the help of equation 2.

$$g_{mean} = \frac{\sum_{i=1}^{row} \sum_{j=1}^{column} g(i,j)}{(row \times column)} \quad \ldots \ldots \ldots (2)$$

If the gray value of any pixel (i, j) (say g(i,j)) is greater than or equal to $g_{mean}$, then the pixel location in the binary image (i, j) is set with 1 (white) else it is set with 0 (black). The binarization process can be mathematically expressed with the help of equation 3.

$$b(i,j) = 1 \text{ if } g(i,j) \geq g_{mean} \\ = 0 \text{ otherwise} \quad \ldots \ldots \ldots (3)$$

In a binary image, black pixels mean background and are represented with '0's, whereas and white pixels mean the face region and are represented with '1's. The binary image corresponding to the grayscale image of Fig. 2b) is shown in Fig. 2c).

## 2.3. Extraction of largest component

- The foreground of a binary image may contain more than one objects or components. Say, in fig. 2c), it has three objects or components. The large one represents the face region. The others are at the left hand bottom corner and a small dot on the top. Then largest component has been extracted from binary image using "Connected Component Labeling" algorithm [13]. This algorithm based on either "4-conneted" neighbours or "8-connected" neighbours method [14]. In "4-connected" neighbours method, a pixel is considered as connected if it has neighbours on the same row or column. This is illustrated in Fig. 4a). Suppose the central pixel of a 3×3 mask "f" is f(x, y), then this method will consider the pixels f(x+1,y), f(x-1,y), f(x,y+1) and f(x,y-1) for checking the connectivity of f(x,y). In "8-connected" method besides the row and columns neighbours, the diagonal neighbours are also checked. That means "4-connected" pixels plus the diagonal pixels are called an "8-connected" neighbour which is illustrated in Fig. 4b). Thus for a central f(x,y) of a 3×3 mask "f" the "8-connected" neighbour methods will consider f(x-1,y-1), f(x-1,y), f(x-1,y+1), f(x,y-1), f(x,y+1), f(x+1,y-1), f(x+1,y) and f(x+1,y+1) for checking the connectivity of f(x,y).

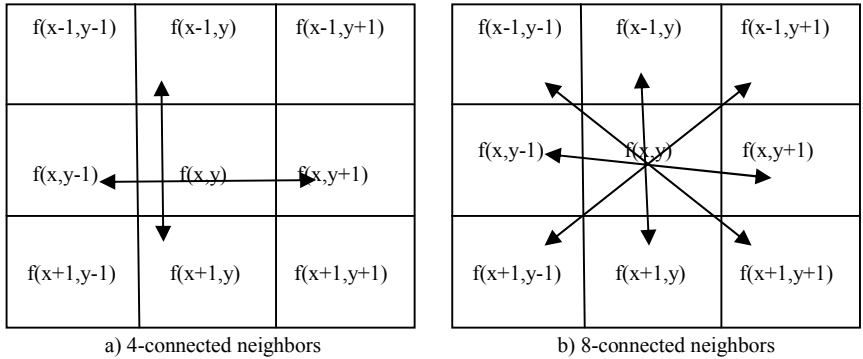

a) 4-connected neighbors         b) 8-connected neighbors

Fig. 3 Different connected neighborhoods

Fig. 4a) Connected component         Fig. 4b) Labeled connected components

"Connected component labeling" algorithm is given below:-

**LabelConnectedComponent(im)**
// LabelConnectedComponent(im) is a method which takes //one argument that is an image named f. f(x,y) is the current pixel on $x^{th}$ row and $y^{th}$ column.

1. Consider the whole image pixel by pixel in row wise manner in order to get connected component. Let f(x,y) be the current pixel position
   **if** ( f(x,y)==1) and f(x,y) does not any labelled neighbour in its 8-connected neighbourhood) **then**
      Create a new label for f(x,y).
   **else**
      **if** (f(x,y) has only labelled neighbour) **then**
         Mark f(x,y) with that label.
   e**lse**
      **if** (f(x,y) has two or more labeled) **then**
         Choose one of the labels for f(x,y) and memorize that these labels are equivalent.
2. Go another pass through the image to determine the equivalencies and labeling each pixel with a unique label for its equivalence class.

Using the above "Connected component labeling" algorithm, the largest component of face region is identified from Fig. 2c) which is shown in Fig. 5.

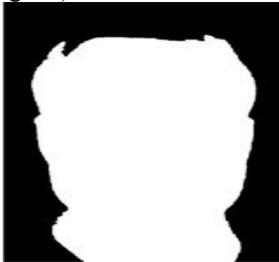    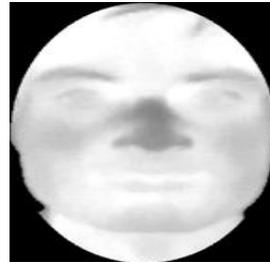

Fig. 5 A largest components as a face skin region        Fig. 6 Cropped face region in elliptic shape

*2.4. Finding the centroid [15]*

Centroid has been extracted from the largest component of the binary image using equation 4 and 5.

$$X = \frac{\sum m_{f(x,y)} x}{\sum m_{f(x,y)}} \quad \ldots \ldots \ldots (4) \qquad Y = \frac{\sum m_{f(x,y)} y}{\sum m_{f(x,y)}} \quad \ldots \ldots \ldots (5)$$

Where x, y is the co-ordinate of the binary image and m is the intensity value that is $m_{f(x, y)}$=f(x, y)=0 or 1.

*2.5. Cropping of the face region in elliptic shape*

Normally human face is of ellipse shape. Then from the above centroid co-ordinates, human face has been cropped in elliptic shape using "Bresenham ellipse drawing" [16] algorithm. This algorithm takes the distance between the centroid and the right ear as the minor axis of the ellipse and distance between the centroid and the fore head as major axis of the ellipse. The pixels selected by the ellipse drawing algorithm are mapped onto the gray level image of Fig. 2b) and finally the face region is cropped. This is shown in Fig. 6.

*2.6. Calculate LL and HL/LH/HH average band using Haar wavelet transform*

The first method of feature extraction is discrete wavelet transform (DWT). The DWT was invented by the Hungarian mathematician Alfréd Haar in 1909. A key advantage of wavelet transform over Fourier transforms is temporal resolution. Wavelet transform captures both frequency and spatial information. The DWT has a huge number of applications in science, engineering, computer science and mathematics.

The Haar transformation is used here since it is the simplest wavelet transform of all and can successfully serve our purpose. Wavelet transform has merits of multi-resolution, multi-scale decomposition, and so on. To obtain the standard decomposition [17] of a 2D image, the 1D wavelet transform to each row is applied first. This operation gives an average pixel value along with detail coefficients for each row. These transformed rows are treated as if they were themselves in an image. Now, 1D wavelet transform to each column is applied. The resulting pixel values are all detail coefficients except for a single overall average coefficient. As a result the elliptical shape facial image is decomposed into four regions can be gained. These regions are one low-frequency $LL_1$ (approximate component), and three high-frequency regions (detailed components), namely $LH_1$ (horizontal component), $HL_1$ (vertical component), and $HH_1$ (diagonal component), respectively. The low frequency sub-band $LL_1$ can be further decomposed into four sub-bands $LL_2$, $LH_2$, $HL_2$ and $HH_2$ at the next coarse scale. $LL_i$ is a reduced resolution corresponding to the low-frequency part of an image. The sketch map of the quadratic wavelet decomposition is shown in Fig. 6.

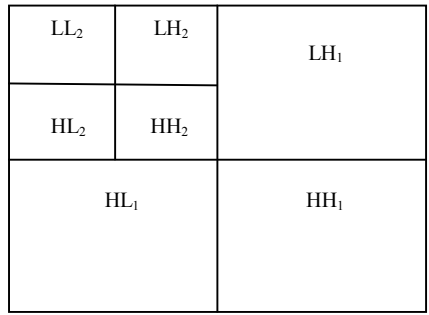

Fig. 7 Sketch map of the quadratic wavelet decomposition.

As illustrated in Fig. 7, the L denotes low frequency and the H denotes high frequency, and that subscripts named from 1 to 2 denote simple, quadratic wavelet decomposition respectively. The standard decomposition algorithm is given below:-

**function StandardDecomposition(Im[1:r,1:c])**
// Im[1:r,1:c] is an image realized by 2D array, where r is the number of rows and c is the number of column.
    **for** i=1:r
        1D wavelet transform (Im(i,:))
    **end**
    **for** j=1:c
        1D wavelet transform (Im(:,j))
    **end**
**end**

Let's start with a simple example of 1D wavelet transform [18]. Suppose an image with only one row of four pixels, having intensity values [10 4 9 5]. Now apply the Haar wavelet transform on this image. To do so, first pair up the input intensity values or pixel values, storing the mean in order to get the new lower resolution image with intensity values [7 7]. Obviously, some information might be lost in this averaging process. Some detail coefficients need to be stored to recover the original four intensity values from the two mean values, which capture the missing information. In this example, 3 is the first detail coefficient, since the computed mean is 3 less than 10 and 3 more than 4. This single number is responsible to recover the first two pixels of original four-pixel image. Similarly, the second detail coefficient is 2. Thus, the original image is decomposed into a lower resolution (two-pixel) version and a pair of detail coefficients. Repeating this process recursively on the averages gives the full decomposition, which is shown in Table 2:

Table 2. Resolution, mean and the detail coefficients of full decomposition.

| Resolution | Mean | Detail coefficients |
|---|---|---|
| 4 | [10 4 9 5] | |
| 2 | [7 7] | [3 2] |
| 1 | [7] | [0] |

Thus, for the one-dimensional Haar transform of the original four-pixel image is given by [7 0 3 2]. After applying standard decomposition algorithm on Fig. 6, the resultant figure is shown in Fig. 8.

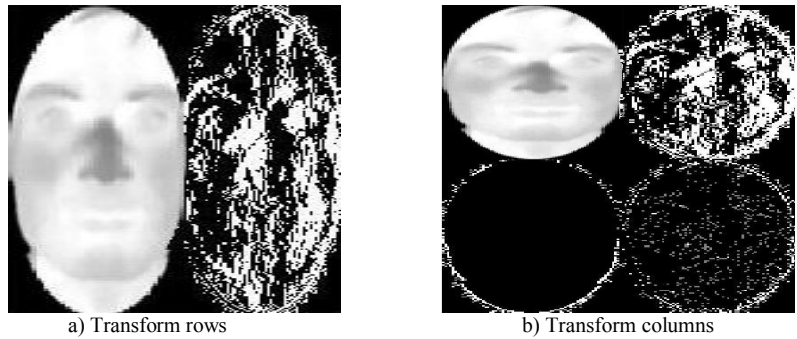

a) Transform rows          b) Transform columns

Fig. 8 Haar Wavelet Transform

The pixels of $LL_2$ image can be rearranged horizontally or vertically. So the image can be treated as a vector (called as feature vector).

*2.7. Calculation total confidence value*

In the present work wavelet transform is used on the elliptic shape face region once which divide the whole image into 4 equal sized sub-images namely low-frequency LL band (approximate component), and three high frequency bands (detailed components),HL,LH and HH. Then the pixel wise average of the detail components is computed using equation 6.

$$D(x,y) = \frac{1}{3}(A(x,y) + B(x,y) + C(x,y)) \ldots \ldots \ldots (6)$$

Where A(x, y) is the HL band sub-image, B(x, y) is the LH band sub-image and C(x, y) is the HH band sub-image. D(x, y) is the average sub-image of A(x, y), B(x, y) and C(x, y) band sub-images and x, y are spatial co-ordinates.

Next, a matrix called total confidence matrix T(x,y) is formed by taking a pixel-wise weighted sum of pixel values of LL band and average subimages [19-21] using equation 7, as given bellow:

$$T(x,y) = (\alpha(x,y) \times L(x,y)) + (\beta(x,y) \times D(x,y)) \ldots \ldots \ldots (7)$$

Where T(x,y) is the total confidence value, L(x,y) is the LL band subimages and D(x,y) is the average of HL/LH/HH band subimages, while α(x,y) and β(x,y) denotes the weighting factors for pixel values of LL band and HL/LH/HH average band sub-images respectively, which is shown in Fig 9.

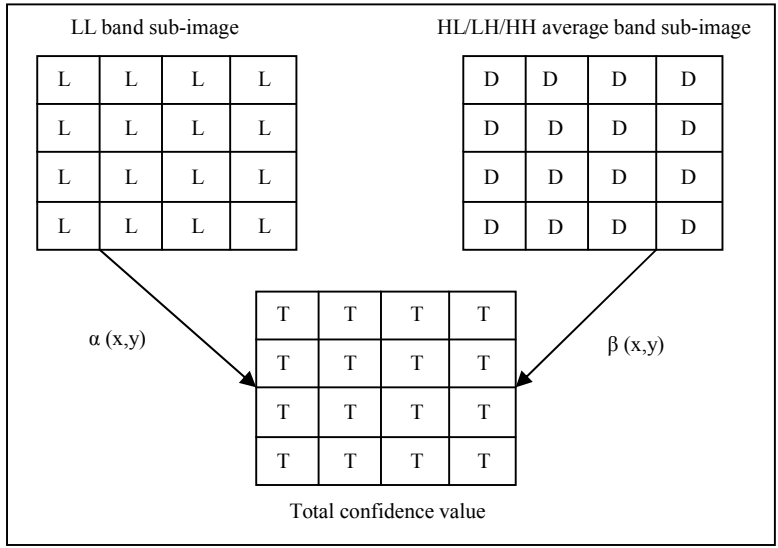

Fig. 9 Mixing technique

After calculating the total confidence matrices for all the images, each matrix is transformed into a horizontal vector, by concatenating the rows of elements in it. This process is repeated for all the images in the database. Let the number of elements in each such horizontal vector be N, where N is the product of the number of rows and columns in LL band or average sub images. By placing the horizontal vectors in row order, a new matrix of size M×N is formed, where M is the number of images in the database. Thus M×N matrix is divided into two parts by the size of (M/2) ×N, of which one part will be used for training purpose and the other part for testing purpose only. The first part contains odd number of images like first row , third row , fifth row  and so on from M×N matrix and the second part contains even number of images like second row , fourth row , sixth row and so on from M×N matrix.

*2.8. Eigenface for recognition*

Principal component analysis (PCA) [22], [23] is performed on training set described above which gives a set of eigenvalues and corresponding eigenvector. Each eigenvector can be shown as sort of ghostly face which is called an eigenface. Each face image in the training set can be represented exactly in terms of a linear combination of these eigenfaces. So, the number of rows i.e. number of face images in the training set is equal to the number of a eigenfaces. However the faces can also be approximated using only the "best" eigenfaces, those that have the largest eigenvalues and which therefore account for the most variance within the set of face images. For this, the eigenvalues are sorted in descending order and eigenvectors corresponding to a few largest eigenvalues are retained. The n-dimensional space is formed by these eigenvector or eigenfaces is called eigenspace. The face images in the training set are then projected on to the eigenspace to get the corresponding eigenfaces, which are then used to train a classifier. For the test face images, similar procedure is followed to get their corresponding eigenfaces, which are classified by the trained classifier.

*2.9. Local Binary Pattern*

The second method of feature is Local Binary Pattern (LBP). The LBP is a type of feature used for texture classification in computer vision. LBP was first described in 1994 [24], [25]. It has since been found to be a powerful feature for texture classification. As it can be appreciated in Fig. 10, the original LBP operator represents each pixel of an image by thresholding its 3x3- neighborhood with the center value and considering the result as a binary number, called the LBP code. In the classification step, the image is usually divided into rectangular regions and histograms of the LBP codes are calculated over each of them. The histograms of each region are concatenated into a single one and a dissimilarity measure is used to compare the histograms of two different images.

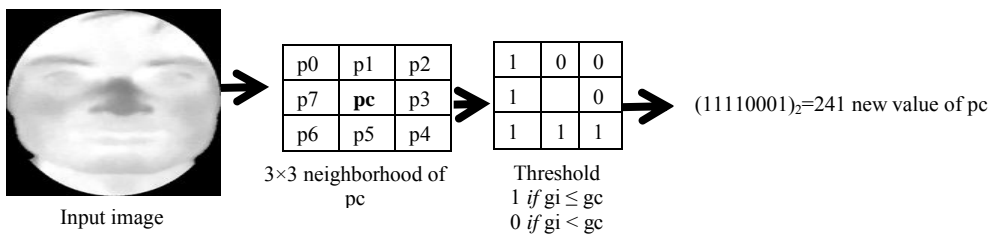

Fig. 10 Local Binary Pattern

*2.10. Multi layer feed forward neural network*

Artificial Neural networks (ANN) [26], [27], posses extraordinary generalization capability to obtain useful information from complex environment or data. So, ANN can be used to extract patterns and detect trends that are too hard to be found by either humans or other computer techniques. A trained ANN can be thought of as an "expert system". The Back propagation learning algorithm is one of the most popular Neural Networks to the scientific and engineering community for modeling and processing of many quantitative phenomena. This learning algorithm is applied to multilayer feed forward networks consisting of processing elements with continuous differentiable activation functions. The five layer feed forward back propagation neural network is used here as a classifier. Momentum allows the network to respond to local gradient and to recent trends in the error surface. The momentum is used to back propagation learning algorithm for making weight changes equal to the sum of a fraction of the last weight change and the new change. The magnitude of the effect that the last weight change is allowed is known as momentum constant (mc). The momentum constant may be any number between 0 and 1. The momentum constant zero means, a weight changes according to the gradient and the momentum constant one means, the new weight change is set to equal the last weight change and the gradient is not considered here. The gradient is computed by summing the gradients calculated at each training example, and the weights and biases are only updated after all training examples have been presented. Tan-sigmoid transfer functions is used to calculate a layer's output from its net the first input and the next three hidden layers and the outer most layer gradient descent with momentum training function is used to updates weight and bias values.

## 2.11. Minimum Distance Classifier

Recogniztion techniques based on matching represent each class by a prototype pattern vector. It places an unknown pattern in the class to which it is closest in terms of a predefined metric. The simplest approach is the minimum distance classifier [15]. It must determine the Euclidean distance between the unknown pattern and each of the prototype vectors. It chooses the smallest distance to take a judgment. The prototype of each pattern class is represented as the mean vector of the patterns of that class which is expressed using equation 8.

$$m_j = \frac{1}{N_j} \sum_{x \in \omega_j} x_j \qquad j = 1,2,3,\ldots,w \qquad \ldots (8)$$

Where W is the number of pattern classes $\omega_j$ is the set of pattern vectors of class j and $N_j$ is the number of pattern vectors in $\omega_j$. In order to get the class membership of an unknown pattern vector x its closest prototype is searched using Euclidean distance measure, which is shown in equation 9.

$$D_j(x) = \|x - m_j\| \qquad j = 1,2,3,\ldots,W \qquad \ldots (9)$$

If $D_j(x)$ is the smallest distance i.e. best match, then assign x to class $\omega_j$.

## 3. Experiment and Results

Experiments have been performed on our own captured thermal face images at our laboratory and Terravic Facial Infrared database. In our database, there are 17×34=578 thermal images. The details of our database have been mentioned in section 2.1. Twelve images are taken in each person for our experiments from two above mentioned datasets, out of which 6 face images are used to form training set and 6 face images are used to form testing set. We have made all the images of size 112×92. The "Terravic Facial Infrared database" contains total no. of 20 classes (19 men and 1 woman) of 8-bit gray scale JPEG thermal faces of 320×240. Size of the database is 298MB and images with different rotations are left, right and frontal face images also available with different items like glass and hat [13]. Experimental process can be divided into several ways:

## 3.1. Harr wavelet + PCA + ANN

In the first set of experiments Haar wavelet is used to decompose the cropped face image once which produces 4 sub-images as LL, HL, LH and HH bands. Then the average of HL/LH/HH band sub-images is computed using equation (6). We have used ten different sets of values for (α, β) to generate 10 different confidence matrices for each face image. The values of α and β are chosen according to equation 10.

$$\left.\begin{array}{l}\beta = 0.1 \times i \\ \alpha = 1.0 - \beta\end{array}\right\} \quad 0 \leq i \leq 10 \quad \ldots\ldots\ldots\ldots (10)$$

After computing the confidence matrices of all the decomposed face images, PCA is performed on these confidence matrices for further dimensionality reduction. ANN classifier (with 0.02 acceleration and 0.9 momentum) is then used to classify the face images on the basis of the extracted features. The recognition performances of the classifier on our own database and Terravic Facial IR database are shown in Tables 3 and 4 respectively. The results are also shown graphically in Figures 11 and 12 respectively.

Table 3 Recognition performance (own database) with varying numbers of eigenvectors and the values of α and β.

| Value of α and β | Recognition rate (%) | | | | |
|---|---|---|---|---|---|
| | 10 eigen vectors | 20 eigen vectors | 30 eigen vectors | 40 eigen vectors | 50 eigen vectors |
| α=1.0 and β=0.0 | 88.23 | 91.18 | 91.18 | 88.23 | 83.33 |
| α=0.9 and β=0.1 | 86.27 | 83.33 | 91.18 | 91.18 | 83.33 |
| α=0.8 and β=0.2 | 81.38 | 88.23 | 88.23 | **95.09** | 74.50 |
| α=0.7 and β=0.3 | 81.38 | 78.57 | 76.74 | 86.27 | 81.38 |
| α=0.6 and β=0.4 | 88.23 | 91.18 | 86.27 | 88.23 | 78.57 |
| α=0.5 and β=0.5 | 94.11 | 78.57 | 91.18 | 83.33 | 91.18 |
| α=0.4 and β=0.6 | 86.27 | 88.23 | 78.57 | 86.27 | 86.27 |
| α=0.3 and β=0.7 | 83.33 | 83.33 | 83.33 | 86.27 | 88.23 |
| α=0.2 and β=0.8 | 81.38 | 83.33 | 76.74 | 88.23 | 83.33 |
| α=0.1 and β=0.9 | 83.33 | 88.23 | 83.33 | 81.38 | 86.27 |
| α=0.0 and β=1.0 | 86.27 | 78.57 | 76.74 | 76.74 | 81.38 |

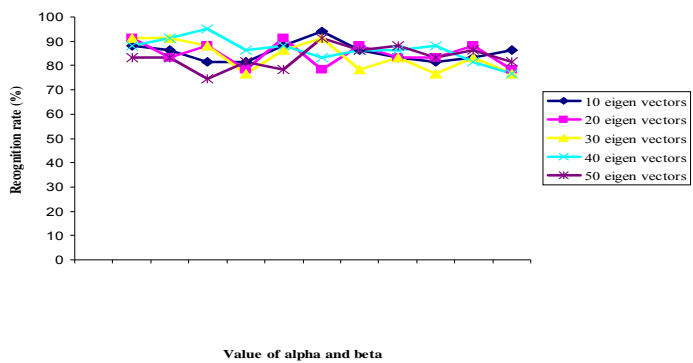

Fig. 11 Comparative study of recognition performance (own database) with varying numbers of eigenvectors and values of α and β

Table 4 Recognition performance (benchmark database) with varying numbers of eigenvectors and the values of α and β.

| Value of α and β | Recognition rate (%) | | | | |
|---|---|---|---|---|---|
| | 10 eigen vectors | 20 eigen vectors | 30 eigen vectors | 40 eigen vectors | 50 eigen vectors |
| α=1.0 and β=0.0 | 83.33 | 86.27 | 80.39 | 86.27 | 75.49 |
| α=0.9 and β=0.1 | 86.27 | 86.27 | 75.49 | 89.22 | 86.27 |
| α=0.8 and β=0.2 | 89.22 | 83.33 | 89.22 | 83.33 | 80.39 |
| α=0.7 and β=0.3 | 89.22 | 83.33 | 86.27 | 80.39 | 83.33 |
| α=0.6 and β=0.4 | 86.27 | 80.39 | 86.27 | 89.22 | 89.22 |
| α=0.5 and β=0.5 | 94.11 | 83.33 | 92.15 | 89.22 | 89.22 |
| α=0.4 and β=0.6 | 80.39 | 78.57 | 86.27 | 92.15 | 88.22 |
| α=0.3 and β=0.7 | 80.39 | 86.27 | 89.22 | 86.27 | 83.33 |
| α=0.2 and β=0.8 | 89.22 | 83.33 | 86.27 | 80.39 | 78.57 |
| α=0.1 and β=0.9 | 89.22 | 78.57 | 80.39 | 89.22 | 83.33 |
| α=0.0 and β=1.0 | 80.39 | 80.39 | 83.33 | 75.49 | 83.33 |

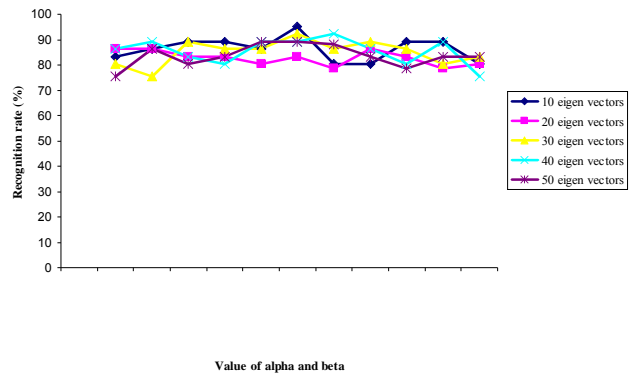

Fig. 12 Comparative study of recognition rate (performed on Terrivic Facial thermal database) with varying numbers of eigenvectors and values of α and β

*3.2. Harr wavelet + PCA + Minimum distance classifier*

In the second set of experiments, the feature set was kept the same as those in the first set of experiments, but the classifier is chosen as minimum distance classifier. The recognition performance obtained on both the thermal face databases considered here are detailed in Tables 5 and also graphically compared in Fig. 13.

15Table 5 Recognition performance (on own database and benchmark database) with minimum distance classifier and the value of α and β.

| Value of α and β | Recognition rate (%) | |
|---|---|---|
| | Own database | Terravic Facial thermal database |
| α=1.0 and β=0.0 | 89.22 | 94.11 |
| α=0.9 and β=0.1 | 86.27 | 86.27 |
| α=0.8 and β=0.2 | 86.27 | 83.33 |
| α=0.7 and β=0.3 | 83.33 | 86.27 |
| α=0.6 and β=0.4 | 86.27 | 83.33 |
| α=0.5 and β=0.5 | 80.39 | 80.39 |
| α=0.4 and β=0.6 | 80.39 | 78.57 |
| α=0.3 and β=0.7 | 80.39 | 86.27 |
| α=0.2 and β=0.8 | 83.33 | 86.27 |
| α=0.1 and β=0.9 | 83.33 | 86.27 |
| α=0.0 and β=1.0 | 86.27 | 80.39 |

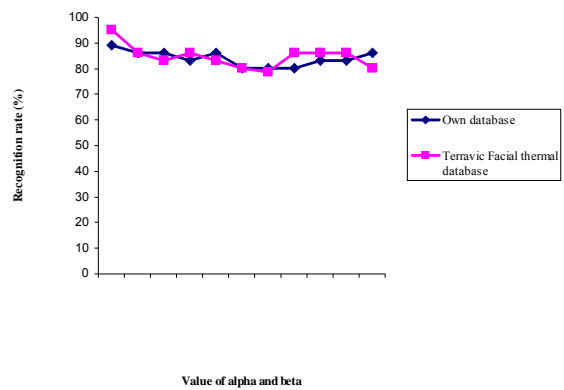

Fig. 13 Comparative study of Recognition performance (own database and benchmark database) with minimum distance classifier and the values of α and β

### 3.3. Local Binary Pattern + (PCA + ANN / Minimum distance classifier)

In the third set of experiments, cropped face images is divided in to 161 subimages each of size 8×8 pixels. Then Local Binary Pattern is used to extract features from each of the subimages which are concatenated in row wise manner. After performing PCA on the LBP features for dimensionality reduction, ANN and minimum distance classifier are used separately for, recognition of the face images on the basis of the extracted features. The obtained recognition results are shown in Table 6. The results are also shown graphically in Fig. 14.

Table 6 Recognition performance (own database and benchmark database) with varying numbers of eigenvectors, ANN and minimum distance classifier.

| | Recognition Rate (%) | | | | | |
|---|---|---|---|---|---|---|
| | ANN classifier | | | | | Minimum distance classifier |
| | 10 eigen vectors | 20 eigen vectors | 30 eigen vectors | 40 eigen vectors | 50 eigen vectors | |
| Own database | 86.27 | 83.33 | 86.27 | 86.27 | 83.33 | 89.22 |
| Terravic Facial thermal database | 86.27 | 89.22 | 83.33 | 92.15 | 89.22 | 94.11 |

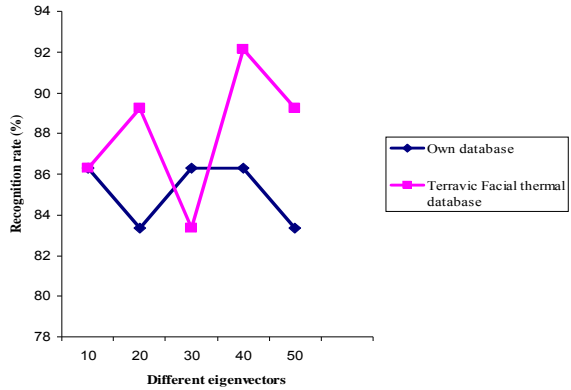

Fig. 14 Recognition performance (own database and benchmark database) with varying numbers of eigenvectors, ANN and minimum distance classifier.

## 4. Conclusions

In this paper a comparative study of thermal face recognition methods is discussed and implemented. In this study two local-matching techniques, one based on Haar wavelet and the other based on Local Binary Pattern are analyzed. Firstly, human thermal face images are pre-processed and cropped the face region only, from the entire face images. Then above mentioned two feature extraction methods are used to extract features from the cropped images. Then PCA is performed on the individual feature set for dimensionality reducation. Finally two different classifiers are used to classify face images. One such classifier is multi-layer feed forward neural network and another is minimum distance classifier. The Experiments have been performed on the database created at our own laboratory and Terravic Facial IR Database. The proposed system gave higher recognition performance in the experiments and the recognition rate was 95.09% for α=0.8, β=0.2 and number of eigen vectors is 40. This experiment was performed on our own database, which is shown in Table 3. Furthermore, no knowledge of geometry or specific feature of the face is required. However, this system is applicable to front views and constant background only. It may fail in unconstraint environments like natural scenes.

### Acknowledgements

Authors are thankful to a major project entitled "Design and Development of Facial Thermogram Technology for Biometric Security System," funded by University Grants Commission (UGC),India and "DST-PURSE Programme" at Department of Computer Science and Engineering, Jadavpur University, India for providing necessary infrastructure to conduct experiments relating to this work.